\documentclass[10pt,twocolumn,letterpaper]{article}

\usepackage{cvpr}
\usepackage{times}
\usepackage{epsfig}
\usepackage{graphicx}
\usepackage{amsmath,bm}
\usepackage{amssymb}
\usepackage{hyperref}       % hyperlinks
\usepackage{url}            % simple URL typesetting
\usepackage{booktabs}       % professional-quality tables
\usepackage{amsfonts}       % blackboard math symbols
\usepackage{nicefrac}       % compact symbols for 1/2, etc.
\usepackage{microtype}      % microtypography
\hypersetup{colorlinks=true, linkcolor=red}
\usepackage{pgfplots}
\usepackage{pgfplotstable}
\usepackage{filecontents,pgfplots}
\usepackage{tikz}
\usetikzlibrary{arrows,calc,shapes,snakes,positioning,matrix,arrows,decorations.pathmorphing,decorations.text}

\usepackage{mathrsfs}
\usepackage{soul}
\usepackage{multirow}
\usepackage{comment}
\usepackage{rotating}
\usepackage{mwe}
\PassOptionsToPackage{bookmarks=false}{hyperref}
\cvprfinalcopy % *** Uncomment this line for the final submission

\def\cvprPaperID{2503} % *** Enter the CVPR Paper ID here

\DeclareMathAlphabet\mathbfcal{OMS}{cmsy}{b}{n}

\begin{document}

%%%%%%%%% TITLE
\title{Adversarially Learned One-Class Classifier for Novelty Detection}

\author{Mohammad Sabokrou$^1$, Mohammad Khalooei$^2$, Mahmood Fathy$^1$, %Juan Carlos Niebles$^4$, 
Ehsan Adeli$^3$\\
$^1$Institute for Research in Fundamental Sciences\\$^2$Amirkabir University of Technology\quad$^3$Stanford University
% \and
}

\maketitle
%\thispagestyle{empty}

%%%%%%%%% ABSTRACT
\begin{abstract}
Novelty detection is the process of identifying the observation(s) that differ in some respect from the training observations (the target class). In reality, the novelty class is often absent during training, poorly sampled or not well defined. Therefore, one-class classifiers can efficiently model such problems. However, due to the unavailability of data from the novelty class, training an end-to-end deep network is a cumbersome task. In this paper, inspired by the success of generative adversarial networks for training deep models in unsupervised and semi-supervised settings, we propose an end-to-end architecture for one-class classification. Our architecture is composed of two deep networks, each of which trained by competing with each other while collaborating to understand the underlying concept in the target class, and then classify the testing samples. One network works as the novelty detector, while the other supports it by enhancing the inlier samples and distorting the outliers. The intuition is that the separability of the enhanced inliers and distorted outliers is much better than deciding on the original samples. The proposed framework applies to different related applications of anomaly and outlier detection in images and videos. The results on MNIST and Caltech-256 image datasets, along with the challenging UCSD Ped2 dataset for video anomaly detection illustrate that our proposed method learns the target class effectively and is superior to the baseline and state-of-the-art methods. 
\end{abstract}

%%%%%%%%% BODY TEXTn
\section{Introduction}
Novelty detection is the process of identifying the new or unexplained set of data to determine if they are within the norm (\ie, inliers) or outside of it (\ie, outliers). Novelty refers to the unusual, new observations that do not occur regularly or is simply different from the others. Such problems are especially of great interest in computer vision studies, as they are closely related to outlier detection \cite{xia2015learning,you2017provable},  image denoising \cite{buades2005non}, anomaly detection in images \cite{cong2011sparse,li2014anomaly} and videos \cite{sabokrou2017deep}. Novelty detection can be portrayed in the context of one-class classification \cite{moya1996network,gardner2006one,khan2014one}, which aims to build classification models when the negative class is absent, poorly sampled or not well defined. As such, the negative class can be considered as the novelty (\ie, outlier or anomaly), while the positive (or target) class is well characterized by instances in the training data. 

\begin{figure}[t]
%\begin{center}
%   \includegraphics[width=\linewidth]{images/fig1.pdf}
  (a)  \raisebox{-.5\height}{\includegraphics[width=0.95\linewidth]{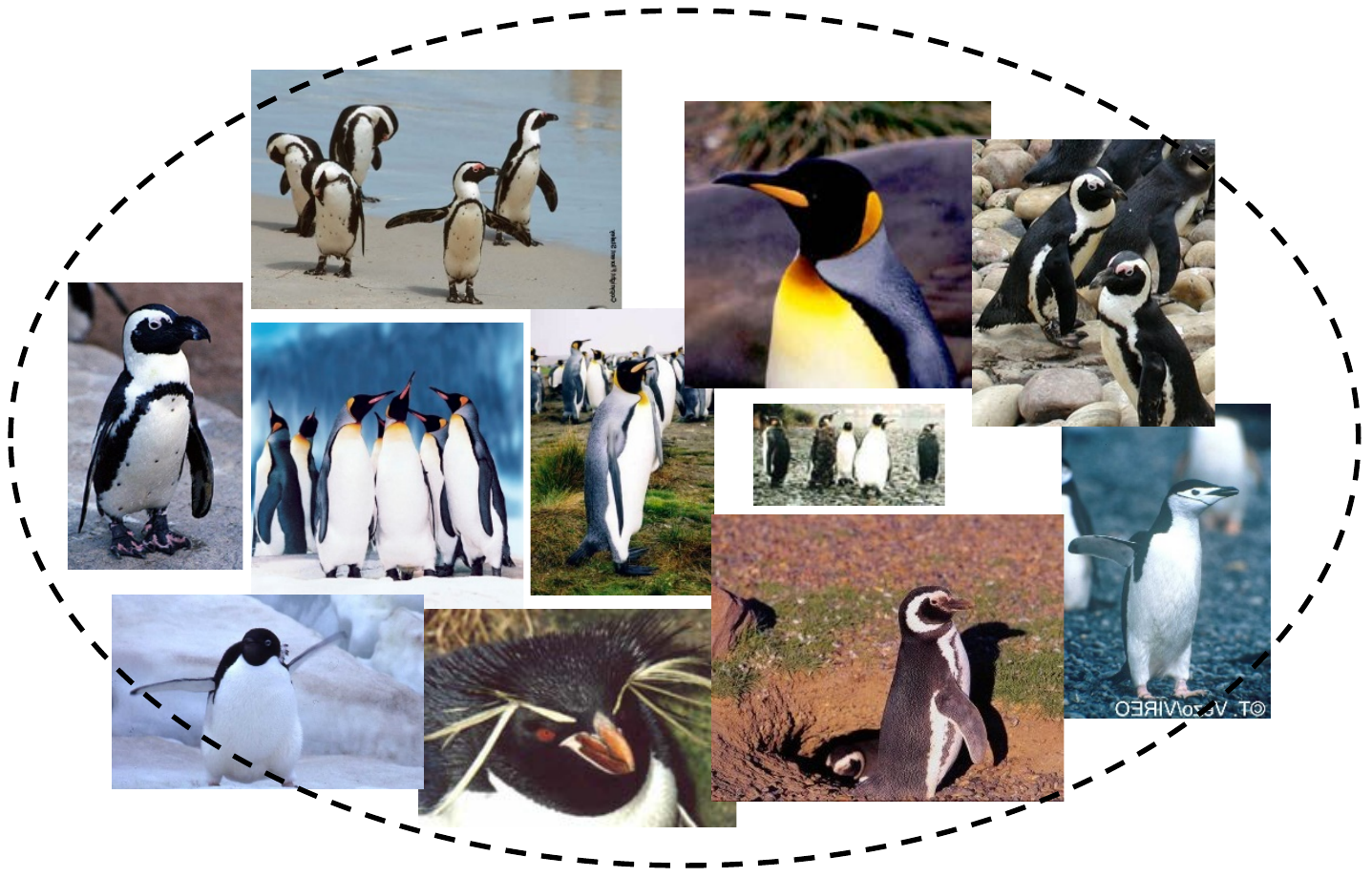}}
\addtolength{\tabcolsep}{-5pt}    
% \begin{tabular}{cccccc}
%     & \multicolumn{2}{c}{Inlier Samples} &  & \multicolumn{2}{c}{Outlier Samples} \\
% {\footnotesize $X$} & \raisebox{-.5\height}{\includegraphics[width=0.18\linewidth]{images/(3).jpg}} & \raisebox{-.5\height}{\includegraphics[width=0.18\linewidth]{images/(12).jpg}} & ~~ & \raisebox{-.5\height}{\includegraphics[width=0.18\linewidth]{images/(1).png}} & \raisebox{-.5\height}{\includegraphics[width=0.18\linewidth]{images/(2).png}} \\
% {\footnotesize $\tilde{X}=X+\eta$} & \raisebox{-.5\height}{\includegraphics[width=0.18\linewidth]{images/(5).jpg}} & \raisebox{-.5\height}{\includegraphics[width=0.18\linewidth]{images/(4).png}} & ~~ & \raisebox{-.5\height}{\includegraphics[width=0.18\linewidth]{images/(6).png}} & \raisebox{-.5\height}{\includegraphics[width=0.18\linewidth]{images/(7).png}} \\
% {\footnotesize $\mathcal{R}(\tilde{X})$} & \raisebox{-.5\height}{\includegraphics[width=0.18\linewidth]{images/(11).png}} & \raisebox{-.5\height}{\includegraphics[width=0.18\linewidth]{images/(12).png}} & ~~ & \raisebox{-.5\height}{\includegraphics[width=0.18\linewidth]{images/(8).png}} & \raisebox{-.5\height}{\includegraphics[width=0.18\linewidth]{images/(9).jpg}} \\
% {\footnotesize $\mathcal{D}(\tilde{X})$} & 0.75 & 0.72 & ~ & 0.53 & 0.27 \\
% {\footnotesize $\mathcal{D}(\mathcal{R}(\tilde{X}))$} & {\bf 0.85} & {\bf 0.91} & ~~ & 0.25 & 0.10 \\
% \end{tabular}
\\ (b) \begin{tabular}{cccccc}
    & \multicolumn{2}{c}{Noisy Inlier Samples} &  & \multicolumn{2}{c}{Outlier Samples} \\
{\footnotesize $X$} & \raisebox{-.5\height}{\includegraphics[width=0.18\linewidth]{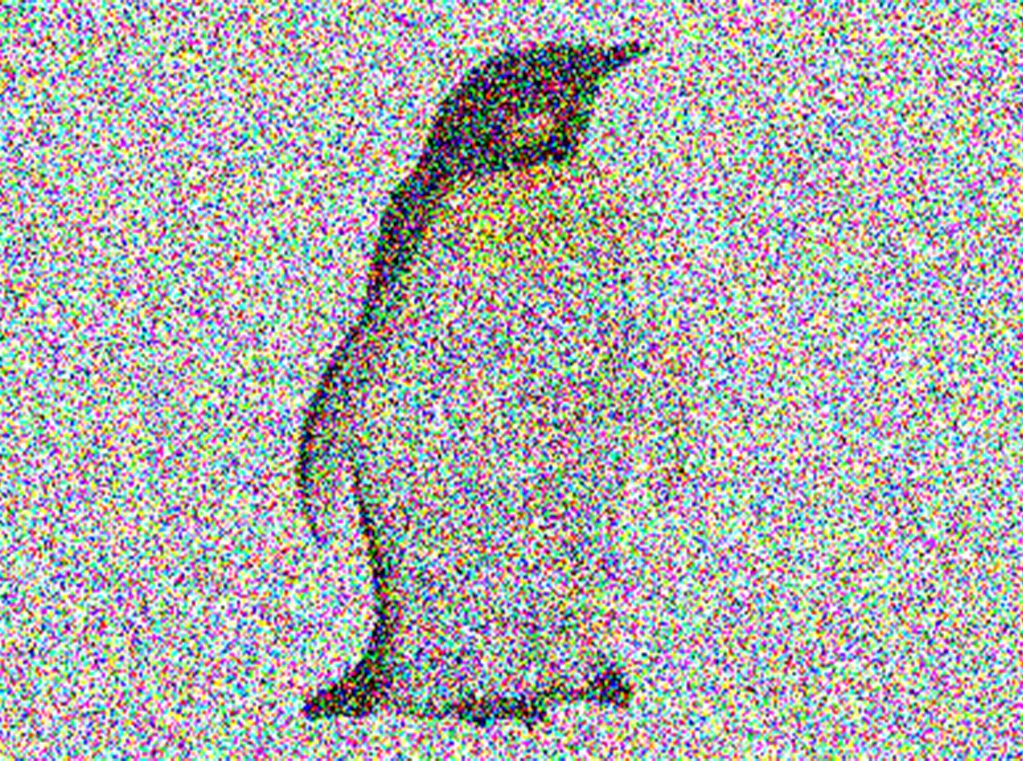}} & \raisebox{-.5\height}{\includegraphics[width=0.18\linewidth]{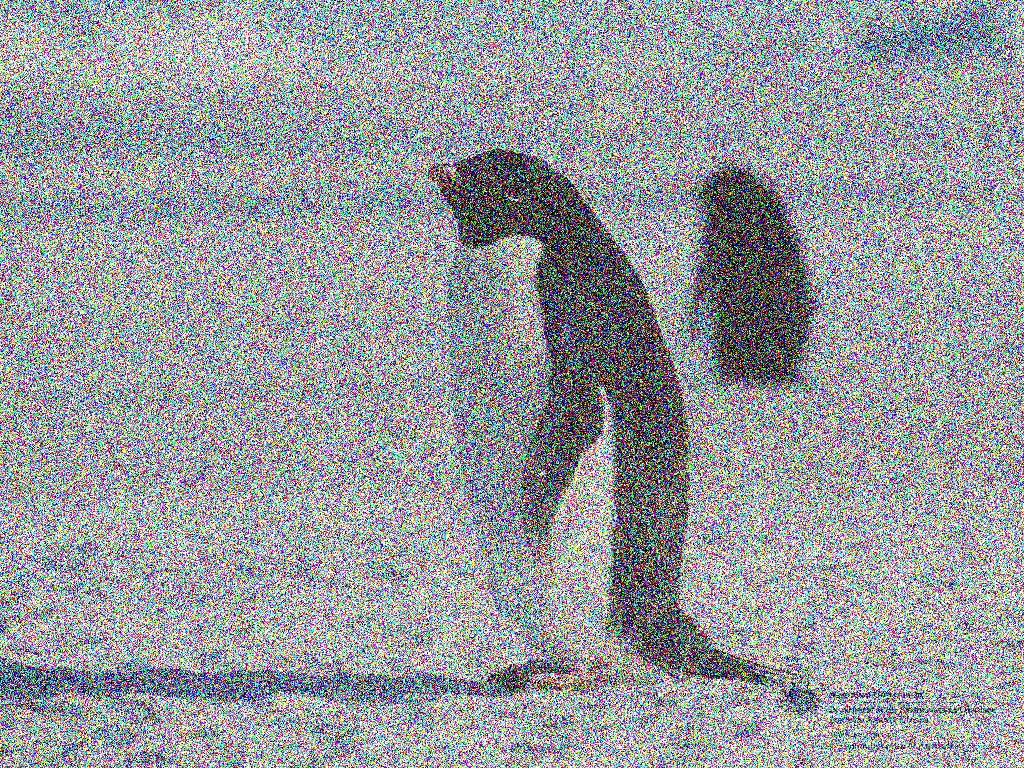}} & ~~ & \raisebox{-.5\height}{\includegraphics[width=0.18\linewidth]{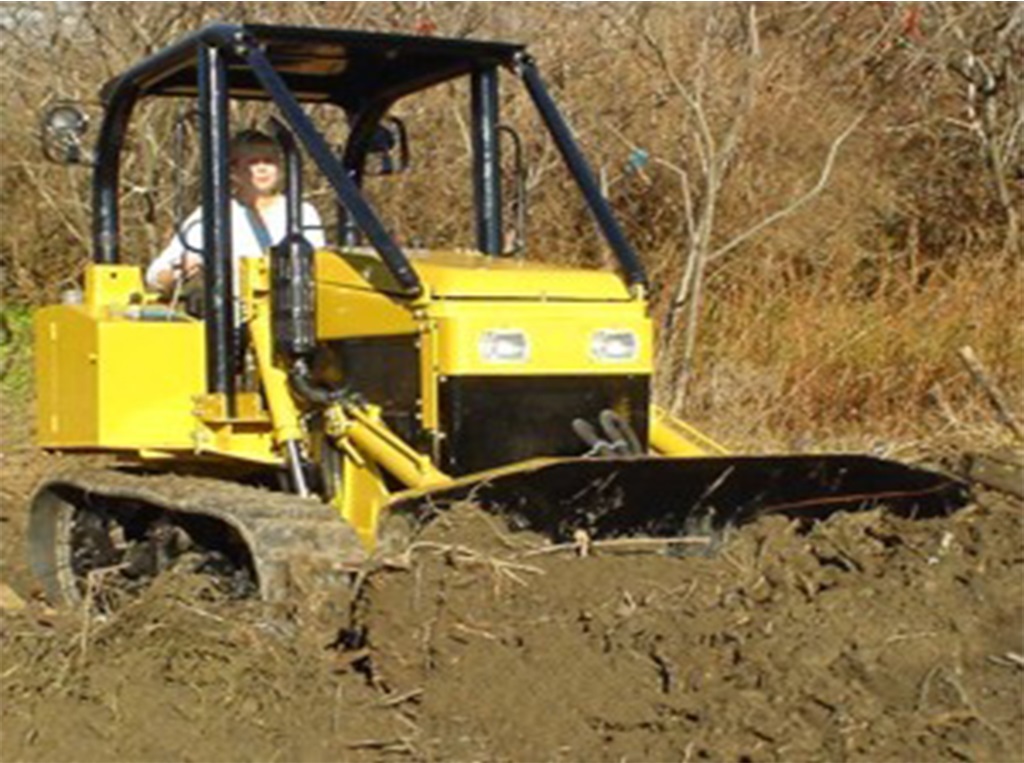}} & \raisebox{-.5\height}{\includegraphics[width=0.18\linewidth]{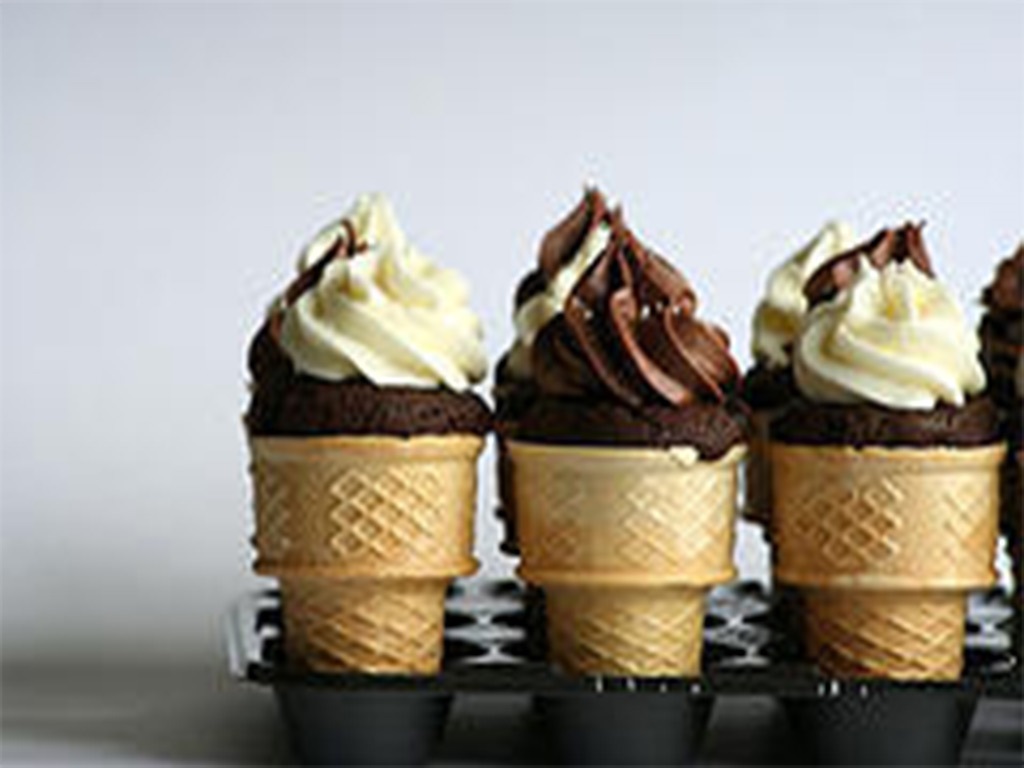}} \\
{\footnotesize $\mathcal{R}(X)$} & \raisebox{-.5\height}{\includegraphics[width=0.18\linewidth]{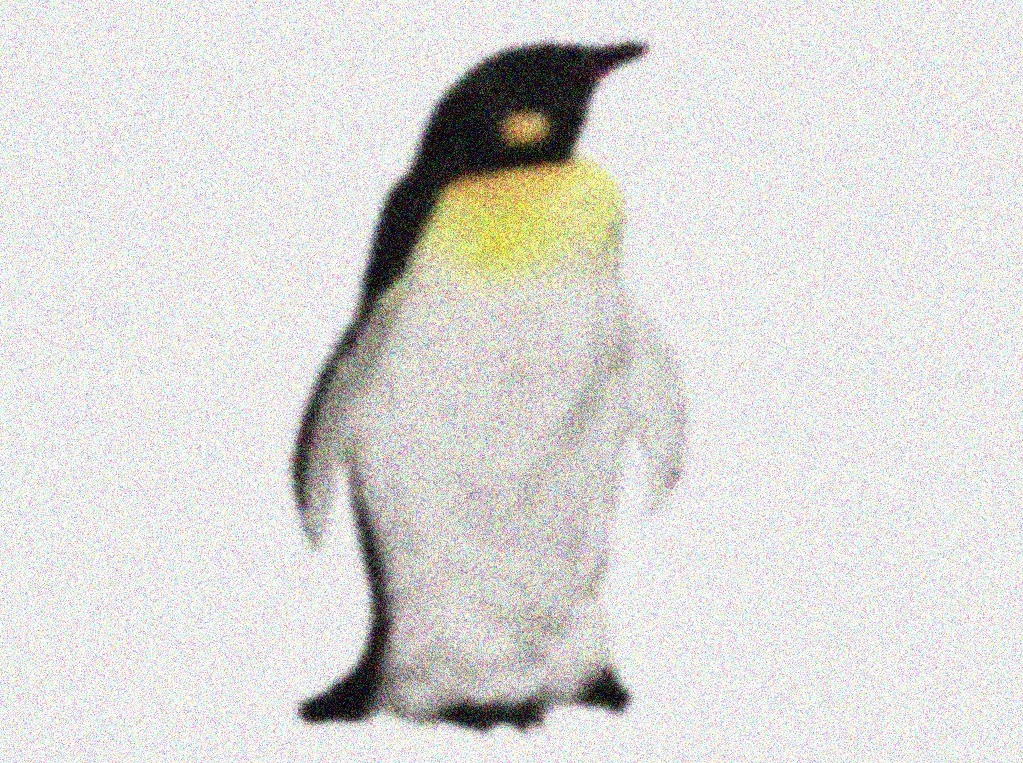}} & \raisebox{-.5\height}{\includegraphics[width=0.18\linewidth]{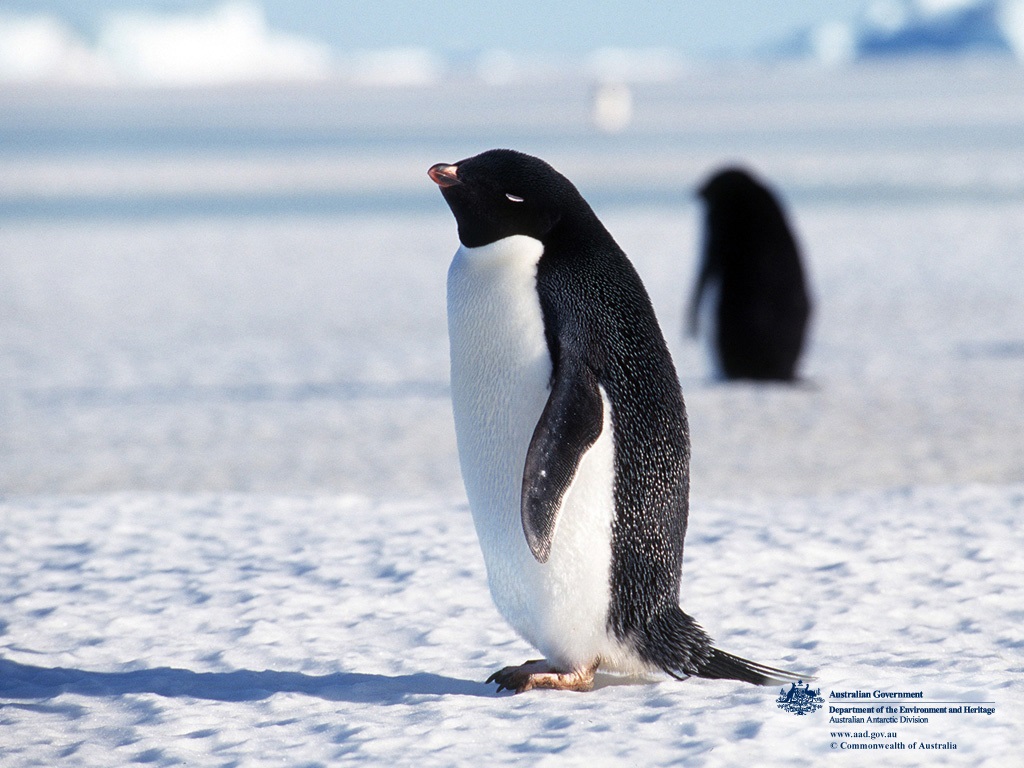}} & ~~ & \raisebox{-.5\height}{\includegraphics[width=0.18\linewidth]{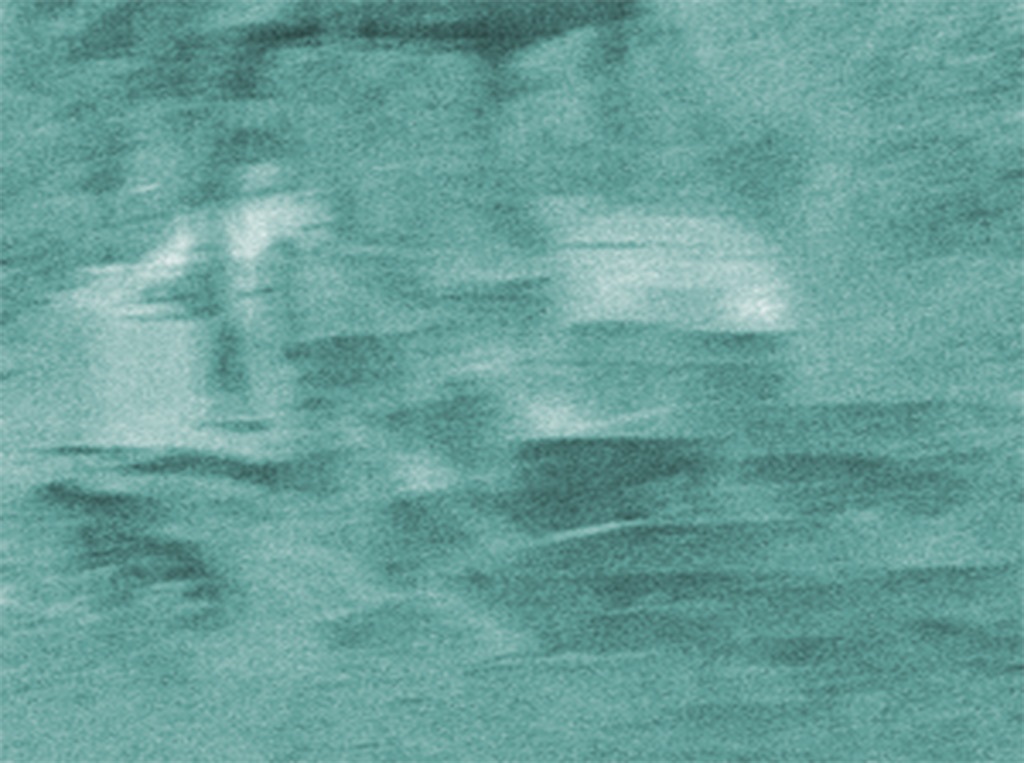}} & \raisebox{-.5\height}{\includegraphics[width=0.18\linewidth]{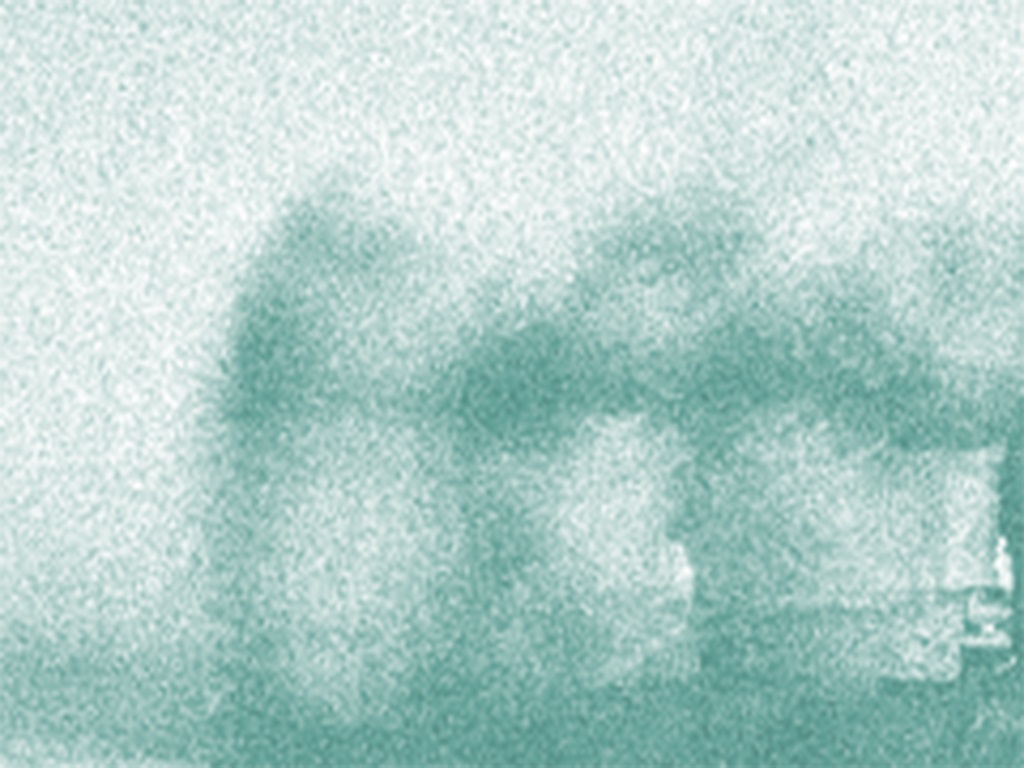}} \\
{\footnotesize $\mathcal{D}({X})$} & 0.75 & 0.72 & ~ & 0.53 & 0.27 \\
{\footnotesize $\mathcal{D}(\mathcal{R}({X}))$} & {\bf 0.85} & {\bf 0.91} & ~~ & 0.25 & 0.10 \\
\end{tabular}
\addtolength{\tabcolsep}{5pt}    
%\end{center}
   \caption{Example outputs of the proposed model, trained to detect Penguins (a), in response to inlier and outlier samples (b). The first row of (b) shows some example images, and the second row contains the output of the $\mathcal{R}$ network on them, \ie, $\mathcal{R}(X)$. As can be seen, $\mathcal{R}$ enhanced the inlier samples (even in the presence of noise) but distorted the outliers. Two last rows show the score of $\mathcal{D}$ applied to $X$ and $\mathcal{R}({X})$, respectively. $\mathcal{R}({X})$ is indeed more separable than only using the original input image, $X$.}
%   \caption{Example outputs of the proposed model, trained to detect Penguins, in response to inlier and outlier samples. First row shows some example original samples, which are contaminated by some Gaussian noise (\ie, $\eta$), shown in the second row (denoted by $\tilde{X}$). The third row contains the output of the $\mathcal{R}$ network on the noisy image,  $\mathcal{R}(\tilde{X})$. As can be seen, $\mathcal{R}$ enhanced the inlier samples, but destructed the outlier ones. Two last rows indicates the score of $\mathcal{D}$ with $\tilde{X}$ as its input, and $\mathcal{R}(\tilde{X})$, respectively. $\mathcal{R}(\tilde{X})$ is indeed more separable than using the input image, $\tilde{X}$.}
 \label{fig:fig1}
\end{figure}

To accurately chart the intrinsic geometry of the positive class, the first step is to efficiently represent the data in a way that can entangle more or less the different explanatory factors of variation in the data. Recently, deep learning approaches have gained immense success in representing visual data for various vision-based applications \cite{simonyan2014very,szegedy2015going}, especially in cases that they are trained in an end-to-end fashion. However, for novelty detection or one-class classification applications, due to unavailability of data from the negative class, training an end-to-end deep network is not straightforward. Some efforts have been made, in recent years, to benefit from deep features in learning one-class classifiers \cite{xu2015learning,sabokrou2017deep,ravanbakhsh2017abnormal,lawson2017finding,sabokrou2018deep,sabokrou2017fast}, few of which could train an end-to-end feature learning and classification model.

Inspired by the recent developments in generative adversarial networks (GANs) \cite{goodfellow2014generative}, we propose an end-to-end model for one-class classification and apply it to different applications including outlier detection, novelty detection in images and anomaly event detection in videos. The proposed architecture, similar to GANs, comprises two modules, which compete to learn while collaborating with each other for the detection task. The first module (denoted as $\mathcal{R}$) refines the input and gradually injects discriminative material into the learning process to make the positive and novelty samples (\ie, inliers, and outliers) more separable for the detector, the second module (referred to as $\mathcal{D}$).

These two networks are adversarially and unsupervisedly learned using the training data, which is composed of only the target class. Specifically, $\mathcal{R}$ learns to reconstruct the positive samples and tries to fool the detector (\ie, $\mathcal{D}$). Whereas, $\mathcal{D}$ learns to distinguish original (positive) samples from the reconstructed ones. In this way, $\mathcal{D}$ learns merely the concept characterized by the space of all positive samples, and hence it can be used for distinguishing between positive and novelty classes. On the other hand, $\mathcal{R}$ learns to efficiently reconstruct the positive samples, while for negative (or novelty) samples it is unable to reconstruct the input accurately, and hence, for negative samples it acts as a decimator (or informally a distorter). In the testing phase, $\mathcal{D}$ operates as the actual novelty detector, while $\mathcal{R}$ improves the performance of the detector by adequately  reconstructing the positive or target samples and decimating (or distorting) any given negative or novelty samples. Fig.  \ref{fig:fig1} depicts example inputs and outputs of both $\mathcal{R}$ and $\mathcal{D}$ networks for a model trained to detect images of Penguins.

In summary, the main contributions of this paper are as follows: (1) We propose an end-to-end deep network for learning one-class classifier learning. To the best of our knowledge, this article is one of the firsts to introduce an end-to-end network for one-class classification. (2) Almost all approaches based on GANs in the literature \cite{radford2015unsupervised} discard either the generator or the discriminator (analogous to $\mathcal{R}$ and $\mathcal{D}$ِ, respectively, in our architecture) after training. Only one of the trained models is used, while our setting is more efficient and benefits from both trained modules to collaborate in the testing stage. (3) Our architecture learns the model in the complete absence of any training samples from the novelty class and achieves state-of-the-art performance in different applications, such as outlier detection in images and anomaly event detection in videos.

%The rest of this paper is organized as follow, ....

%-------------------------------------------------------------------------
\section{Related Works}
\label{sec:rw}
One-class classification is closely related to rare event detection, outlier detection/removal, and anomaly detection. All these applications share the search procedure for a novel concept, which is scarcely seen in the data and hence can all be encompassed by the umbrella term \textit{novelty detection}. Consequently, a wide range of real-world applications can be modeled by one-class classifiers. Conventional research often models the target class, and then rejects samples not following this model. Self-representation \cite{xia2015learning,you2017provable,cong2011sparse,sabokrou2016video} and statistical modeling \cite{markou2003novelty} are the commonly used approaches for this task. For data representation, low level features \cite{bertini2012multi}, high level features (\eg, trajectories \cite{morris2011trajectory}), deeply learned features \cite{xu2015learning,sabokrou2015real,sabokrou2017deep} are used in the literature. A brief review of state-of-the-art novelty detection methods especially the ones based on adversarial learning in deep networks is provided in this section.

\textbf{Self-Representation. }Several previous works show that self-representation is a useful tool for outlier or novelty event detection. For instance, \cite{cong2011sparse,sabokrou2016video} proposed self-representation techniques for video anomaly detection and outlier detection through learning a sparse representation model, as a measure for separating inlier and outlier samples. It is assumed that outlier or novel samples are not sparsely represented using the samples from the target class. In some other works (like \cite{xu2015learning,cong2011sparse}), testing samples are reconstructed using the samples from the target class. The decision if it is an inlier or outlier (novel) is made based on the reconstruction error, \ie, high reconstruction error for a sample indicates that it is more probably an outlier sample. In another work, Liu \etal~\cite{liu2010robust} proposed to use a low-rank self-representation matrix in place of a sparse representation, penalized by the sum of unsquared self-representation errors. This penalization leads to more robustness against outliers (similar to \cite{adeli2015robust}). Similarly, auto-encoders are also exploited to model and measure the reconstruction error for the related tasks of outlier removal and video anomaly detection, in \cite{sabokrou2016video, xu2015learning}.

\textbf{Statistical Modeling. }More conventional methods tend to model the target class using a statistical approach. For instance, after extracting features from each sample, a distribution function is fit on the samples from the target class, and samples far from this distribution are considered as outliers or novelty (\eg,  \cite{eskin2000anomaly,yamanishi2004line,markou2003novelty}). In another work, Rahmani and Atia \cite{rahmani2016coherence} proposed an algorithm termed Coherence Pursuit (CoP) for Robust Principal Component Analysis (RPCA). They assumed that the inlier samples have high correlations and can be spanned in low dimensional subspaces, and hence they have a strong mutual coherence with a large number of data points. As a result, the outliers either do not accord with the low dimensional subspace or form small clusters. Also, a method proposed in \cite{xu2010robust}, OutlierPursuit, used convex optimization techniques to solve the PCA problem with robustness to corrupted entries, which led to the development of many recent methods for PCA with robustness to outliers. Lerman \etal~\cite{lerman2015robust} described a convex optimization problem for detecting the outliers and called it REAPER, which can reliably fit a low-dimensional model to the target class samples. 

\textbf{Deep Adversarial Learning. }In the recent years, GANs \cite{goodfellow2014generative,salimans2016improved} have shown outstanding success in generating data for learning models. They have also been extended to classification models even in the presence of not enough labeled training data (\eg, in \cite{lawson2017finding,schlegl2017unsupervised,ravanbakhsh2017training}). They are based on a two-player game between two different networks, both concurrently trained in an unsupervised fashion. One network is the generator ($G$), which aims at generating realistic data (\eg, images), while the second network poses as the discriminator ($D$), and tries to discriminate real data from the data generated by $G$. One of the different types of GANs, closely related to our work, is the conditional GANs, in which $G$ takes an image $X$ as the input and generates a new image $X'$. Whereas, $D$ tries to distinguish $X$ from $X'$, while $G$ tries to {\it fool} $D$ producing more and more realistic images. Very recently Isola \etal~\cite{isola2016image} proposed an ``Image-to-image translation'' framework based on conditional GANs, where both $G$ and $D$ are conditioned on the real data. They showed that a U-Net encoder-decoder \cite{ronneberger2015u}  with skip connections could be used as the generator coupled with a patch-based discriminator to transform images with respect to different representations. 
In a concurrent work, \cite{ravanbakhsh2017abnormal} proposed to learn the generator as a reconstructor of normal events, and hence if it cannot properly reconstruct a chunk of the input frames, that chunk is considered an anomaly. However, in our work, the first module (\ie, $\mathcal{R}$) not only reconstructs the target class, but it also helps to improve the performance for the model on any given testing image, by refining samples belonging to the target class, and decimating/distorting the anomaly or outlier samples.

%\cite{lawson2017finding,schlegl2017unsupervised,ravanbakhsh2017training} also detect the anomalous by benefit of GAN structure to learn unsupervisedly the normal inlier samples.  
\section{Proposed Approach}
\begin{figure}[t]
\begin{center}
   \includegraphics[width=1\linewidth]{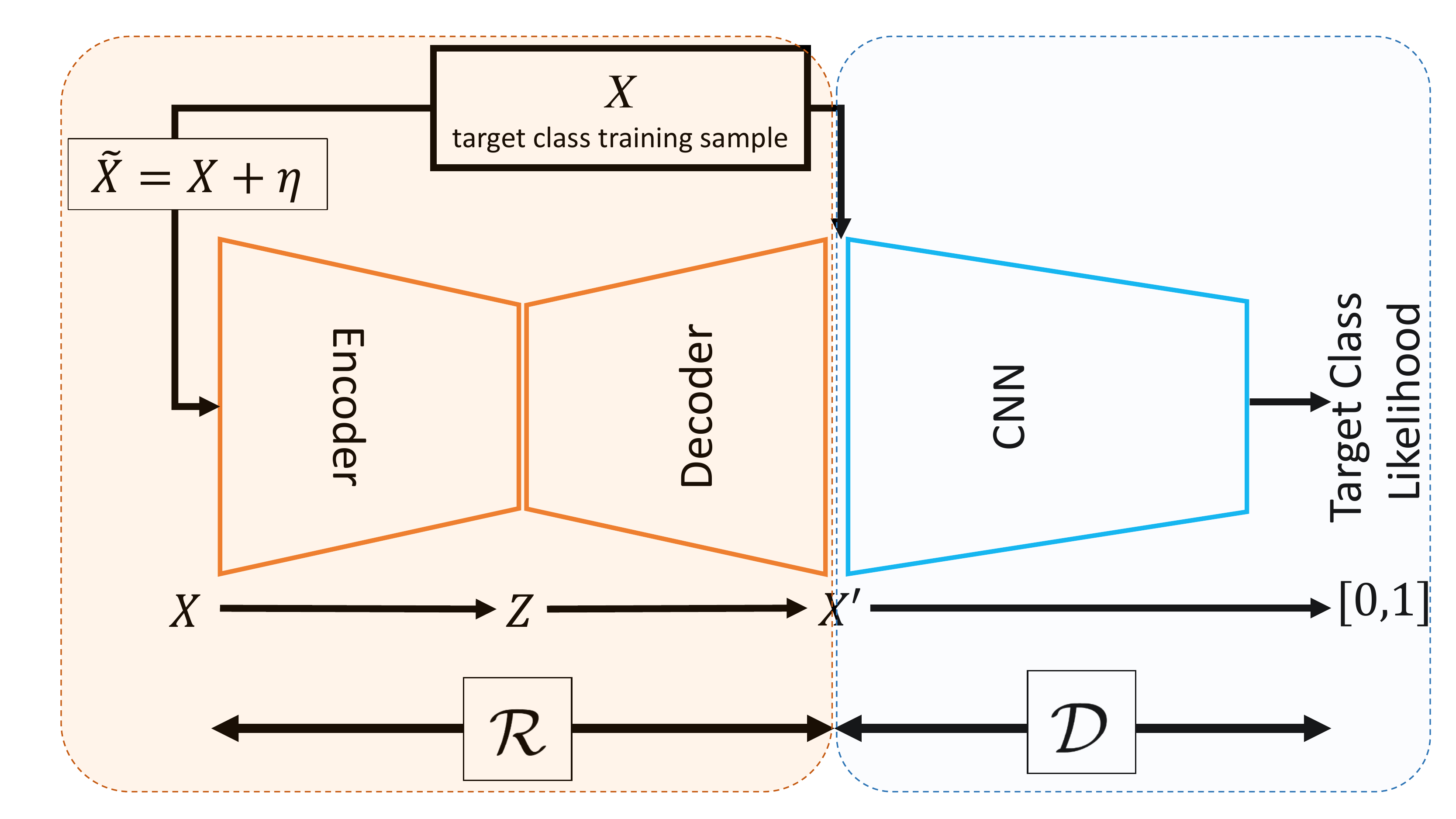}
\end{center}
   \caption{Overview of the proposed structure for one-class classification framework. $\mathcal{R}$ and $\mathcal{D}$ are two modules of the model, which are adversarially learned. $\mathcal{R}$ is optimized to reconstruct samples belonging to the target class, while it works as a decimator function for outlier inputs, whereas $\mathcal{D}$ classifies the input data positive (target) and negative (outlier or anomaly). $\mathcal{D}(\mathcal{R}(X))$ measures the likelihood of the given input sample belonging to the target class.}
\label{fig:onecol}
\end{figure}

The proposed one-class classification framework is composed of two main modules: (1) Network $\mathcal{R}$, and (2) Network $\mathcal{D}$. The former acts as a prepossessing and $\mathcal{R}$efinement (or $\mathcal{R}$econstruction) step, while the latter performs the $\mathcal{D}$iscrimination (or $\mathcal{D}$etection). These two networks are learned in an adversarial and unsupervised manner, within an end-to-end setting. In this section, we present a detailed overview of both. The overall schema of the proposed approach is shown in Fig.  \ref{fig:onecol}. It can be seen that $\mathcal{R}$ reconstructs its input, $X$, generates $X'$, and tries to fool $\mathcal{D}$ so that it speculates that the reconstructed sample is the original data, not a reconstructed sample. On the other hand, $\mathcal{D}$ has access to the original set of data and is familiar with their concept. Hence it will reject the reconstructed samples. These two networks play a game, and after the training stage, in which samples from the target class are presented to the model, $\mathcal{R}$ will become an expert to reconstruct the samples from the target class with a minimum error to successfully fool $\mathcal{D}$. The training procedure leads to a pair of networks, $\mathcal{R}$ and $\mathcal{D}$, which both learn the distribution of the target class. These two modules are trained in a GAN-style adversarial learning framework, forming an end-to-end framework for one-class classification for novelty detection. To make the proposed method more robust against noise and corrupt input samples, a Gaussian noise (denoted by $\eta$ in Fig.  \ref{fig:onecol}) is added to the input training samples and fed to $\mathcal{R}$. Detailed descriptions of each module and the overall training/testing procedures are described in the following subsections.

\subsection{$\mathbfcal{R}$ Network Architecture}
\label{sec:R}
It is previously \cite{xia2015learning,sabokrou2016video} investigated that the reconstruction error of an auto-encoder, trained on samples from the target class, is a useful measure for novelty sample detection. Since the auto-encoder is trained to reconstruct target class samples, the reconstruction error for negative (novelty) samples would be high. We use a similar idea, but in contrast, we do not use it for the detection or the discrimination task. Our proposed model uses the reconstructed image to train another network for the discrimination task. 

\begin{figure}[t]
\begin{center}
   \includegraphics[width=1\linewidth]{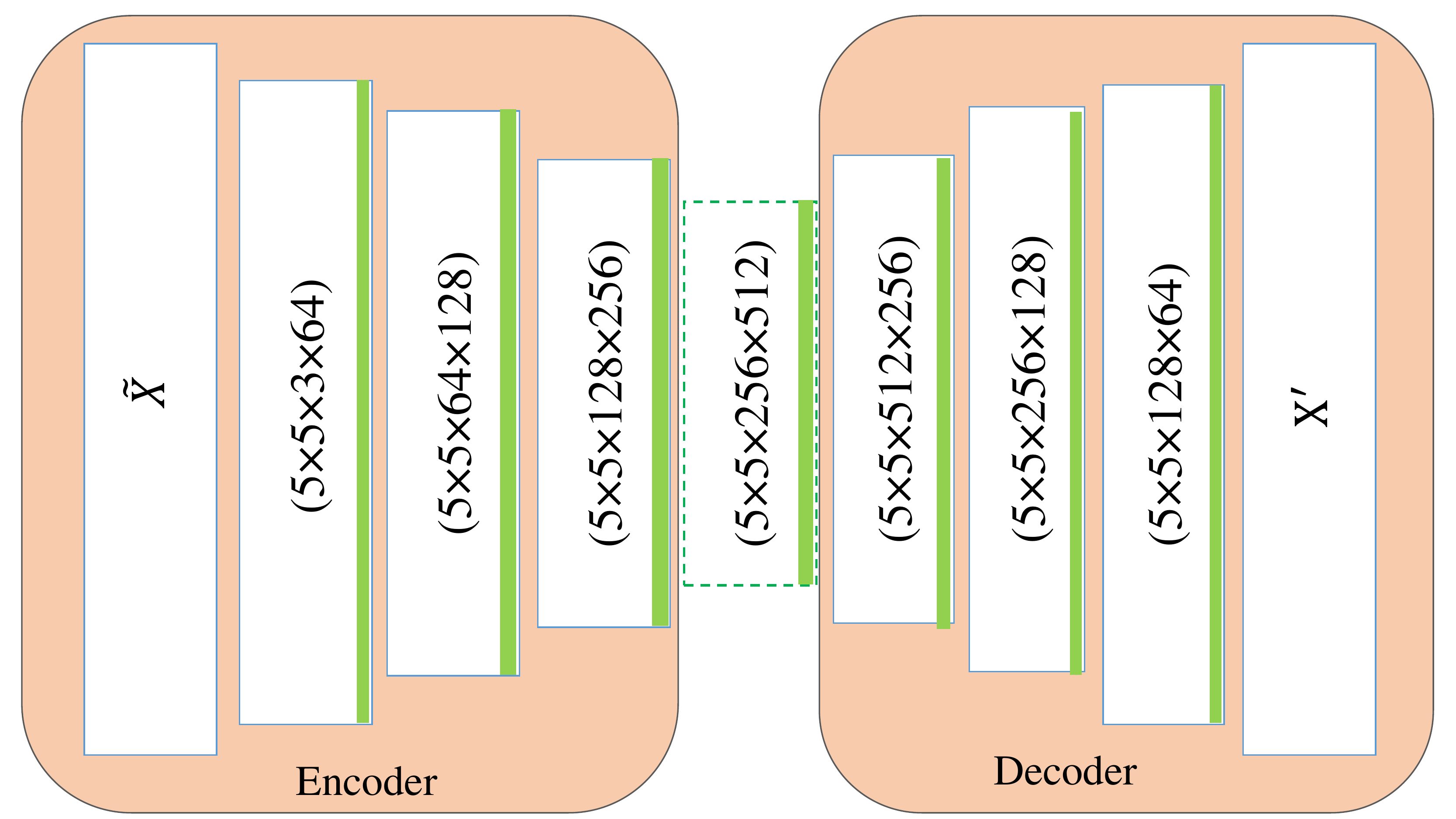}
\end{center}
   \caption{$\mathcal{R}$ network architecture, composed of encoding (first part) and decoding (second part) layers. The properties of each layer are indicated with four hyperparameters in this order: (first dimension of the kernel $\times$ the second dimension of the kernel $\times$ the number of input channels $\times$ the number of output channels).}
\label{fig:R}
\end{figure}

To implement the $\mathcal{R}$ network, we train a decoder-encoder Convolutional Neural Network (CNN) on samples from the target class to map any given input sample to the target concept. As a result, $\mathcal{R}$ will efficiently reconstruct the samples that share a similar concept as the trained target class, while for outlier or novelty inputs, it poorly reconstructs them. In other words, $\mathcal{R}$ enhances the inliers (samples from the target class), while it destructs or decimates the outliers, making it easier for the discriminator to separate the outliers from the vast pool of inliers. Fig.  \ref{fig:R} shows the structure of $\mathcal{R}$ architecture, which includes several convolution layers (as the encoder), followed by several deconvolution layers (for the decoding purpose). For improving the stability of the network similar to \cite{radford2015unsupervised}, we do not use any pooling layers in this network. Eventually, $\mathcal{R}$ learns the concept shared in the target class to reconstruct its inputs based on that concept.  Also, after each convolutional layer, a batch normalization \cite{ioffe2015batch} operation is exploited, which adds stability to our structure. 

 \begin{figure}[b]
\begin{center}
   \includegraphics[width=1\linewidth]{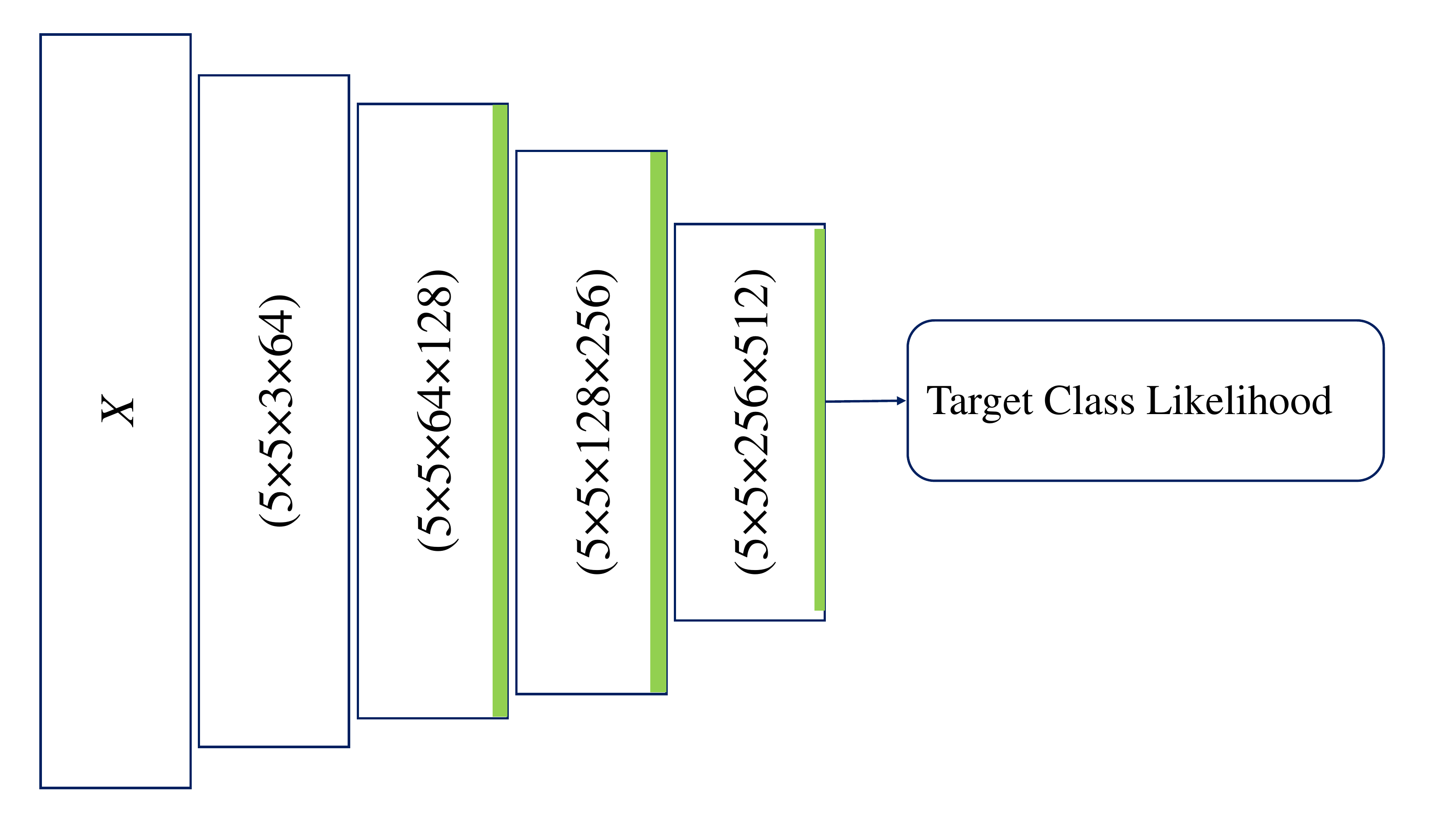}
\end{center}
   \caption{$\mathcal{D}$ network architecture, which determines if its input is from the target class or it is an outlier or novelty. Properties of the layers are denoted similarly to Fig.  \ref{fig:R}.}
\label{fig:D}
\end{figure}
\subsection{$\mathbfcal{D}$ Network Architecture} 
 \label{sec:D}
The architecture for $\mathcal{D}$ is a sequence of convolution layers, which are trained to eventually distinguish the novel or outlier sample, without any supervision. Fig.  \ref{fig:D} shows the details of this network's architecture.  $\mathcal{D}$ outputs a scalar value, relative to the likelihood of its input following the distribution spanned by the target class, denoted by $p_t$. Therefore, its output can be considered as a target likelihood score for any given input. 

%--------------------------------------------------------------------------------------

\subsection{Adversarial Training of $\mathbfcal{R}$+$\mathbfcal{D}$} \label{sec:r+d}
As mentioned in section \ref{sec:rw}, Goodfellow \etal~ \cite{goodfellow2014generative} introduced an efficient way for adversarial learning of two networks (denoted by Generator ($G$) and Discriminator ($D$)), called Generative Adversarial Networks (GANs). GANs aim to generate samples that follow the same distribution as the real data, through adversarial learning of the two networks. $G$ learns to map any random vector, $Z$ from a latent space following a specific distribution, $p_z$, to a data sample that follows the real data distribution ($p_t$ in our case), and $D$ tries to discriminate between actual data and the fake data generated by $G$. Generator and Discriminator are learned in a two-player mini-max game, formulated as: 
\begin{equation}
%\begingroup\makeatletter\def\f@size{7}\check@mathfonts
\begin{aligned}
\min_{G} \max_{D} ~& \Big( \mathbb{E}_{X \sim  p_{t}}[\log(D(X))] \\ & + \mathbb{E}_{Z \sim  p_{z}}[\log(1-D(G(Z)))] \Big).
\end{aligned}
%\endgroup
\end{equation}

In a similar way, we train the $\mathcal{R}$+$\mathcal{D}$ neural networks in an adversarial procedure. In contrast to the conventional GAN, instead of mapping the latent space $Z$ to a data sample with the distribution $p_t$, $\mathcal{R}$ maps
\begin{equation}
\tilde{X} = \left(X \sim  p_t \right) +\left(\eta \sim  \mathcal{N}(0, \sigma^2\mathbf{I}) \right) \longrightarrow X' \sim  p_t,
\end{equation}
in which $\eta$ is the added noise sampled from the normal distribution with standard deviation $\sigma$, $\mathcal{N}(0, \sigma^2\mathbf{I})$. From now on, the noise model is denoted by $\mathcal{N}_\sigma$ for short. This statistical noise is added to input samples to make $\mathcal{R}$ robust to noise and distortions in the input images, in the training stage. As mentioned before, $p_t$ is the supposed distribution of the target class. $\mathcal{D}$ is aware of $p_t$, as it is exposed to the samples from the target class. Therefore, $\mathcal{D}$ explicitly decides if $\mathcal{R}(\tilde{X})$ follows $p_t$ or not. Accordingly, $\mathcal{R}$+$\mathcal{D}$ can be jointly learned by optimizing the following objective: 
\begin{equation}
%\begingroup\makeatletter\def\f@size{7}\check@mathfonts
\begin{aligned}
\min_\mathcal{R} \max_\mathcal{D} ~ & \Big( \mathbb{E}_{X \sim  p_t}[\log(\mathcal{D}(X))] \\
& + \mathbb{E}_{\tilde{X} \sim  p_t+\mathcal{N}_\sigma}[\log(1-\mathcal{D}(\mathcal{R}(\tilde{X})))] \Big), 
\end{aligned}
%\endgroup
\label{eq:q2}
\end{equation}

Based on the above objective (similar to GAN), network $\mathcal{R}$ generates samples with the probability distribution of $p_t$, and as a result its own distribution is given by $p_r\sim \mathcal{R}(X \sim p_t; \theta_{r})$, where $\theta_{r}$ is the parameter of the $\mathcal{R}$ network. Therefore, we want to maximize  $p_t(\mathcal{R}(X \sim p_t ;\theta_{r}))$. 

To train the model, we calculate the loss $\mathcal{L}_{\mathcal{R}+\mathcal{D}}$ as the loss function of the joint network $\mathcal{R}$+$\mathcal{D}$. Besides, we need $\mathcal{R}$'s output to be close to the original input image. As a result, an extra loss is imposed on the output of $\mathcal{R}$:
\begin{equation}
\mathcal{L}_\mathcal{R}=\| X-X'\|^{2}.
\end{equation}

Therefore, the model is optimized to minimize the loss function:
\begin{equation}
\mathcal{L}=\mathcal{L}_{\mathcal{R}+\mathcal{D}}+ \lambda \mathcal{L}_\mathcal{R},
\label{eq:sum_loss}
\end{equation}
where $\lambda > 0$ is a trade-off hyperparameter that controls the relative importance of the two terms. One of the challenging issues for training $\mathcal{R}$+$\mathcal{D}$ is defining an appropriate stopping criterion. Analyzing the loss functions of $\mathcal{R}$ and $\mathcal{D}$ modules to excerpt a stopping criterion based on is a burdensome task, and hence we use a subjective criterion. The training procedure is stopped when $\mathcal{R}$ successfully maps noisy images to clean images carrying the concept of the target class (\ie, favorably fools the $\mathcal{D}$ module). Consequently, we have stopped the training of networks, when $\mathcal{R}$ can  reconstruct its input with minimum error (\ie, $\|X-X'\|^{2} < \rho$, where $\rho$ is a small positive number).

After a joint training of the $\mathcal{R}$+$\mathcal{D}$ network, the behavior of each single one of them can be interpreted as follows:  
\begin{itemize}
\item{$\mathcal{R}(X \sim p_t+\eta) \longrightarrow X'\sim p_t$, where $\| X-X' \|^2$ is minimized. This is because $\theta_{r}$ is optimized to reconstruct those inputs that follow the distribution $p_t$. Note that $\mathcal{R}$ is trained and operates similar to denoising auto-encoders \cite{vincent2008extracting} or, denoising CNNs \cite{divakar2017image}, and its output will be a refined version of the input data. See Figures \ref{fig:fig1} and \ref{fig:recons} for some samples of its reconstructed outputs.}
\begin{figure*}[t]
\begin{center}
   \includegraphics[width=0.8\linewidth]{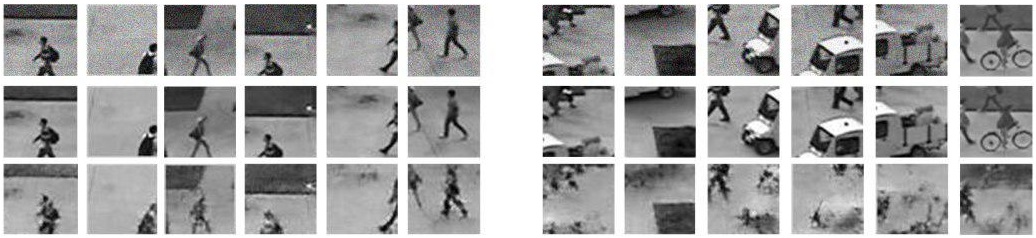}
\end{center}
   \caption{Examples of the output of $\mathcal{R}$ for several inlier and outlier samples from the UCSD Ped2 dataset: $\mathcal{R}$ is learned on normal patches. Left and right samples show the inlier (\ie, target) and outlier (\ie, novelty) samples, respectively. As can be seen, the network $\mathcal{R}$ enhances its input and shows to be robust to the noise present in its input. First row: Patch contaminated by some Gaussian noise; Second row: Original patches; Third row: The output of $\mathcal{R}$ on the noisy samples.}
\label{fig:recons}
\end{figure*}
\item{For any given outlier or novelty sample (denoted by $\hat{X}$) that does not follow $p_t$, $\mathcal{R}$ is confused and maps it into a sample, $\hat{X}'$, with an unknown probability distribution, $p_?$, (\ie, $\mathcal{R}(\hat{X} \nsim p_t + \eta) \longrightarrow \hat{X}' \sim p_?$). In this case, $\|\hat{X}-\hat{X}'\|^2$ cannot become very small or close to zero. This is because $\mathcal{R}$ was not trained on the novelty concept and cannot reconstruct it accordingly (similar to \cite{ravanbakhsh2017abnormal}). Therefore, as a side effect, $\mathcal{R}$ decimates the outliers. As an example, Fig.  \ref{fig:mnist-out} shows samples of a different concept being fed to $\mathcal{R}$ of a network trained to detect digit ``1''.}
\begin{figure}[t]
\begin{center}
   \includegraphics[width=0.7\linewidth]{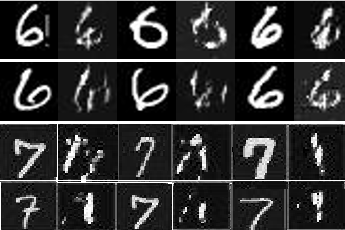}
\end{center}
   \caption{Outputs of $\mathcal{R}$ trained to detect digit ``1'' on MNIST dataset. Samples from classes ``6'' and ``7'' are given to the model as outliers. $\mathcal{R}$ failed to reconstruct them and fundamentally distorted them. In each pair of the images, the first one is the original image and the second one is the output of $\mathcal{R}$.}
\label{fig:mnist-out}
\end{figure}
\item{We can expect that $\mathcal{D}(X' \sim p_t)> \mathcal{D}(\hat{X}' \nsim p_t)$, since $\mathcal{D}$ is trained to detect samples from the distribution $p_t$.} 
\item{It is interesting to note that in most cases $\mathcal{D}(\mathcal{R}(X \sim p_t))-\mathcal{D}(\mathcal{R}(\hat{X} \nsim p_t)) > \mathcal{D}(X \sim p_t)- \mathcal{D}(\hat{X} \nsim p_t)$. This signifies that the output of $\mathcal{R}$ is more separable than original images. It is because of this fact that $\mathcal{R}$ supports $\mathcal{D}$ for better detection. To further explore this, Fig.  \ref{fig:Re} shows the score generated as the output of $\mathcal{D}$ for both cases. In some sensitive applications, it is more appropriate to avoid making decisions on difficult cases \cite{bishop2006pattern}, and leave them for human intervention. These hard-to-decide cases are known to be in the reject region. As shown in Fig. \ref{fig:Re} the reject region of $\mathcal{D}(X)$ is larger than that of $\mathcal{D}(\mathcal{R}(X))$.}
\end{itemize} 
\begin{figure}[t]
\begin{center}
   \includegraphics[width=\linewidth]{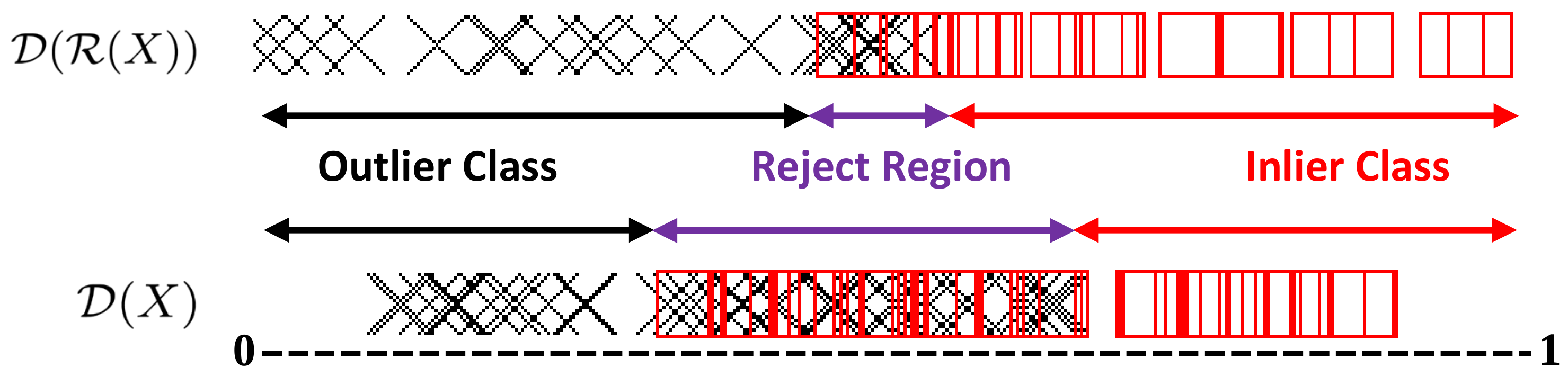}
\end{center}
   \caption{$\mathcal{R}$+$\mathcal{D}$ is trained on inlier samples (digit ``1'') from MNIST dataset. Top: $\mathcal{D}(\mathcal{R}(X))$ scores; Bottom:  $\mathcal{D}(X)$ scores. The scores are generated on 20 inlier and 20 outlier samples. The red squares indicate inlier samples, while $\times$ marks are representatives of the outliers. Reject region for $\mathcal{R}(X)$ is larger than that of $\mathcal{D}(\mathcal{R}(X))$.}
\label{fig:Re}
\end{figure}

\subsection{$\mathbfcal{R}$+$\mathbfcal{D}$: One-Class Classification}
In the previous subsection, characteristics of both $\mathcal{R}$ and $\mathcal{D}$ networks are explained in details. As discussed, $\mathcal{D}$ acts as the novelty detector, and benefits the support of $\mathcal{R}$. Hence, the One-Class Classifier (OCC) can be simply formulated by only using the $\mathcal{D}$ network (similar to \cite{ravanbakhsh2017abnormal}) as:
\begin{equation}
    \text{OCC}_1(X) =  \begin{cases}
    \text{Target Class} & \text{if } \mathcal{D}(X)>\tau, \\
    \text{Novelty (Outlier)} & \text{otherwise},\\
  \end{cases}  
    \label{eq:d}
\end{equation}
where $\tau$ is a predefined threshold. Although the above policy for novelty detection works as well as the state-of-the-art methods (explained in details in the Section \ref{sec:exp}), we further propose to incorporate $\mathcal{R}$ in the testing stage. To this end, $R(X,\theta_{r})$ is used as a refinement step for $X$, in which $\theta_{r}$ is the trained model for the $\mathcal{R}$ module. $\theta_{r}$ is trained to reconstruct and enhance samples that follow $p_t$ (\ie, are from the target class). 
% \begin{equation}
%     R(x) =  \begin{cases}
%     Enhancing      &  \mbox{if} \;\;x \sim  P_{data} \\
%     destructing  & Otherwise\\
%   \end{cases}  
%     \label{eq:t}
% \end{equation}
Consequently, instead of $\mathcal{D}(X)$ we use $\mathcal{D}(\mathcal{R}(X))$. Eq. \eqref{eq:dr} provides a summary of our proposed once-class classification scheme: 
\begin{equation}
    \text{OCC}_2(X) =  \begin{cases}
    \text{Target Class} & \text{if } \mathcal{D}(\mathcal{R}(X)) >\tau, \\
    \text{Novelty (Outlier)} & \text{otherwise}.\\
  \end{cases}  
    \label{eq:dr}
\end{equation}

%-------------------------------------------------------------------------
\section{Experiment Results} \label{sec:exp}
In this section, the proposed method is evaluated on three different image and video datasets. The performance results are analyzed in details and are compared with state-of-the-art techniques. To show the generality and applicability of the proposed framework for a variety of tasks, we test it for detection of \textit{(1) Outlier images}, and \textit{(2) Video anomalies}. 
%Then, we perform several ablation studies on the impact of different aspects and modules of the method.

\subsection{Setup}
All the reported results in this section are from our implementation using the TensorFlow framework \cite{abadi2016tensorflow}, and Python ran on an NVIDIA TITAN X. The detailed structures of $\mathcal{D}$ and $\mathcal{R}$ are explained in details in Sections \ref{sec:D} and \ref{sec:R}, respectively. These structures are kept fixed for different tasks, and $\lambda$ in Eq. \eqref{eq:sum_loss} is set equal to $0.4$. The hyperparameters of batch normalization (as in \cite{ioffe2015batch}) are set as $\epsilon=10^{-6}$ and decay$=0.9$.  

\subsection{Outlier Detection}
As discussed earlier, many computer vision applications face considerable amounts of outliers, since they are common in realistic vision-based training sets. On the other hand, machine learning methods often experience considerable performance degradation in the presence gross outliers, if they fail to deal with processing the data contaminated by noise and outliers. Our method can learn the shared concept among all inlier samples, and hence identify the outliers. Similar to \cite{liu2017incremental,you2017provable,xia2015learning}, we evaluate the performance of our outlier detection method using MNIST\footnote{Available at \url{http://yann.lecun.com/exdb/mnist/}} \cite{lecun2010mnist} and Caltech\footnote{Available at \url{http://www.vision.caltech.edu/Image_Datasets/Caltech256/}} \cite{griffin2007caltech} datasets. 

\subsubsection{Outlier Detection Datasets}

\noindent \textbf{MNIST:} This dataset \cite{lecun2010mnist} includes 60,000 handwritten digits from ``0'' to ``9''. Each of the ten categories of digits is taken as the target class (\ie, inliers), and we simulate outliers by randomly sampling images from other categories with a proportion of $10\%$ to $50\%$. This experiment is repeated for all of the ten digit categories.

\noindent  \textbf{Caltech-256:} This dataset \cite{griffin2007caltech} contains 256 object categories with a total of 30,607 images. Each category has at least 80 images. Similar to previous works \cite{you2017provable}, we repeat the procedure three times and use images from $n \in \{1, 3, 5\}$ randomly chosen categories as inliers (\ie, target). The first 150 images of each category are used, if that category has more than 150 images. A certain number of outliers are randomly selected from the ``clutter'' category, such that each experiment has exactly $50\%$ outliers.

\subsubsection{Outlier Detection Results}
\begin{figure}[h]
\begin{center}
\begin{tikzpicture}
  \begin{axis}[width=8.5cm, height=6cm,
    symbolic x coords = {10, 20, 30, 40, 50},
    legend pos = south west,
    xlabel={Percentage of outliers (\%)},
    ylabel={$F_1$-Score},
    y label style={at={(axis description cs:0.05,.5)}},
  ]
  \addplot+[smooth] coordinates { (10,0.97)(20,0.92)(30,0.92)(40,0.91)(50,0.88)};
  \addplot+[smooth] coordinates { (10,0.93)(20,0.90)(30,0.87)(40,0.84)(50,0.82)};
  \addplot+[smooth] coordinates { (10,0.92)(20,0.83)(30,0.72)(40,0.65)(50,0.55)};
  \addplot+[smooth] coordinates { (10,0.95)(20,0.91)(30,0.88)(40,0.82)(50,0.73)};

  \legend{{\small $\mathcal{D}(\mathcal{R}(X))$ }, {\small $\mathcal{D}(X)$ }, {\small LOF \cite{breunig2000lof}}, {\small DRAE \cite{xia2015learning}}}
  \end{axis}
\end{tikzpicture}
\end{center}
   \caption{Comparisons of $F_1$-scores on MNIST dataset for different percentages of outlier samples involved in the experiment.}
\label{fig:MNIST}
\end{figure}
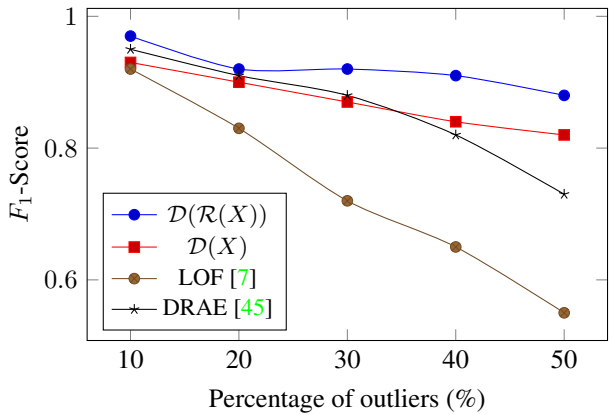

\noindent \textbf{Result on MNIST:} The joint network $\mathcal{R}$+$\mathcal{D}$ is trained on images of the target classes, in the absence of outlier samples. Following \cite{xia2015learning}, we also report the $F_1$-score as a measure to evaluate the performance of our method and compare it with others. Fig.  \ref{fig:MNIST} shows the $F_1$-score of our method (and the state-of-the-art methods) as a function of the portion of outlier samples. As can be seen, our method (\ie, $\mathcal{D}(\mathcal{R}(X))$) operates more efficient than the other two-state-of-the-art methods (LOF \cite{breunig2000lof} and DRAE \cite{xia2015learning}). Also, it is important to note that with the increase in the number of outliers, our method operates consistently robust and successfully detects the outliers, while the baseline methods fail to detect the outliers as their portion increases. Furthermore, one interesting finding of these results is that, based in Fig.  \ref{fig:MNIST}, $\mathcal{D}(X)$ itself can operate successfully well, and outperform the state-of-the-art methods. Nevertheless, it is even improved more when we incorporate $\mathcal{R}$ module, as it modifies the samples (\ie, refines the samples from the target class, and decimates the ones coming from an outlier concept) and helps distinguishability of the samples.

\begin{table*}[t]
\caption{Results on the Caltech-256 dataset: Inliers are taken to be images of one, three, or five randomly chosen categories, and outliers are randomly chosen from category 257-clutter. \textbf{Two first rows:} Inliers are from one category of images, with $50\%$ portion of outliers; \textbf{Two second rows:} Inliers are from three categories of images, with $50\%$ portion of outliers; \textbf{Two last rows:} Inliers come from five categories of images, while outliers compose $50\%$ of the samples. The last two columns have the results or our methods, $\mathcal{D}(X)$ and  $\mathcal{D}(\mathcal{R}(X))$, respectively. Note that in each row the best result is typeset in \textbf{bold} and the second best in \textit{italic} typeface.}
\begin{center}
\begin{tabular}{ccccccccc}
\hline 
  & {\footnotesize CoP \cite{rahmani2016coherence}} &   {\footnotesize REAPER \cite{lerman2015robust}} &  {\footnotesize OutlierPursuit \cite{xu2010robust}} &  {\footnotesize LRR \cite{liu2010robust}} &  {\footnotesize DPCP \cite{tsakiris2015dual}} &  {\footnotesize R-graph \cite{you2017provable}} &   {\footnotesize Ours $\mathcal{D}(X)$ } & {\footnotesize Ours $\mathcal{D}(\mathcal{R}(X))$ }\\
\hline   \hline  
AUC  & 0.905 & 0.816 & 0.837 & 0.907 & 0.783 &\textbf{0.948} & 0.932 & \textit{0.942}\\
$F_1$ & 0.880 & 0.808 & 0.823 & 0.893 & 0.785 &  0.914 & \textit{0.916} & \textbf{0.928}\\
\hline \hline
AUC  & 0.676 & 0.796  & 0.788 & 0.479 & 0.798  & 0.929 & \textit{0.930} & \textbf{0.938}\\
$F_1$  & 0.718 & 0.784 & 0.779  & 0.671 & 0.777 & 0.880 &  \textit{0.902} & \textbf{0.913}\\
\hline \hline
AUC  & 0.487 & 0.657 & 0.629 & 0.337 & 0.676 & \textit{0.913} & \textit{0.913} & \textbf{0.923}\\
$F_1$  & 0.672  & 0.716 &  0.711  & 0.667 & 0.715 &  0.858 &  \textit{0.890} & \textbf{0.905}\\
\hline
\end{tabular}
\end{center}
\label{tab:caltec}
%NOT%\vspace{-2mm}
\end{table*}

\noindent \textbf{Result on Caltech-256:} In this experiment, similar setup as in \cite{you2017provable} is used, and we compare our method with \cite{you2017provable} and 6 other methods therein designed specifically for detecting outliers, including Coherence Pursuit (CoP) \cite{rahmani2016coherence}, OutlierPursuit \cite{xu2010robust}, REAPER \cite{lerman2015robust}, Dual Principal Component Pursuit (DPCP) \cite{tsakiris2015dual}, Low-Rank Representation (LRR) \cite{liu2010robust}, OutRank \cite{moonesignhe2006outlier}. The results are listed in Table \ref{tab:caltec}, which are comprised of $F_1$-score and area under the ROC curve (AUC). The results of other methods are borrowed from \cite{you2017provable}. This table confirms that our proposed method performs at least as well as others, while in many cases it is superior to them. As can be seen, both proposed methods (\ie, $\mathcal{D}(X)$ and  $\mathcal{D}(\mathcal{R}(X))$) outperform all other methods in most cases. Interestingly, as we increase the number of inlier classes, from 1 to 3 and 5 (first, second and the last two rows, respectively), our method robustly learns the inlier concept(s).

\begin{figure}[t!]
\begin{center}
\addtolength{\tabcolsep}{-5pt}    
\begin{tabular}{ccccccc}
    & \multicolumn{3}{c}{Normal Patches} &  & \multicolumn{2}{c}{Anomaly Patches} \\
{\footnotesize $X$} & \raisebox{-.5\height}{\includegraphics[width=0.15\linewidth]{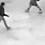}} &
\raisebox{-.5\height}{\includegraphics[width=0.15\linewidth]{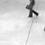}} &
\raisebox{-.5\height}{\includegraphics[width=0.15\linewidth]{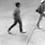}} &~~ & \raisebox{-.5\height}{\includegraphics[width=0.15\linewidth]{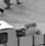}} &  \raisebox{-.5\height}{\includegraphics[width=0.15\linewidth]{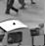}} \\
{\footnotesize $\mathcal{R}(X)$} & \raisebox{-.5\height}{\includegraphics[width=0.15\linewidth]{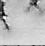}} & \raisebox{-.5\height}{\includegraphics[width=0.15\linewidth]{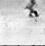}} &  \raisebox{-.5\height}{\includegraphics[width=0.15\linewidth]{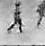}} & ~~ & \raisebox{-.5\height}{\includegraphics[width=0.15\linewidth]{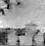}} &
\raisebox{-.5\height}{\includegraphics[width=0.15\linewidth]{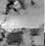}}\\
{\footnotesize $\mathcal{D}({X})$} & 0.15 & 0.19 & 0.32 & ~ & 0.35 & 0.44 \\
{\footnotesize $\mathcal{D}(\mathcal{R}({X}))$} & {\bf 0.44} & {\bf 0.64}&\textbf{0.56} & ~~ & 0.20 & 0.30 \\
\end{tabular}
\addtolength{\tabcolsep}{5pt}    
\end{center}
   \caption{Examples of patches (denoted by $X$) and their reconstructed versions using $\mathcal{R}$ (\ie, $\mathcal{R}(X)$): Three left Columns are normal patches, and two right ones are abnormal. The output of $\mathcal{D}$ is the likelihood of being a normal patch (a scalar in range [0,1]).}
 \label{fig:ped2}
\end{figure}
 
\subsection{Video Anomaly Detection}
Anomaly event detection in videos or visual analysis of suspicious events is a topic of great importance in different computer vision applications. Due to the increased complexity of video processing, detecting abnormal events (\ie, anomaly or novelty events) in videos is even a more burdensome task than image outlier detection. We run our method on a popular video dataset, UCSD \cite{chan2008ucsd} Ped2. The results are reported on a frame-level basis, as we aligned our experimental setup to previous works for comparison purposes. 

\subsubsection{Anomaly Detection Dataset} 

\noindent \textbf{UCSD dataset:} This dataset \cite{chan2008ucsd} includes two subsets, \textbf{Ped1} and \textbf{Ped2}, from two different outdoor scenes, recorded with a static camera at 10 fps and resolutions $158\times 234$ and $240 \times 360$, respectively. The dominant mobile objects in these scenes are pedestrians. Therefore, all other objects (\eg, cars, skateboarders, wheelchairs, or bicycles) are considered as anomalies. We evaluate our algorithm on Ped2.

\subsubsection{Anomaly Detection Results}
\noindent \textbf{Result on UCSD Ped2:} For this experiment, we divide the frames of the video into 2D patches of size $30 \times 30$. Training patches are only composed of frames with normal behaviors. In the testing phase, test patches are given to the joint network $\mathcal{R}$+$\mathcal{D}$, and the results are recorded. Fig.  \ref{fig:ped2} shows examples of the output of $\mathcal{R}$ on the testing patches. As it is evident, normal patches (\ie, left part of the figure) are successfully refined and reconstructed by the $\mathcal{R}$ network, while the abnormal ones (\ie, the right part of the figure) are distorted and not adequately reconstructed. The last two rows in the figure show the likelihood score identified by our methods ($\mathcal{D}(X)$ and $\mathcal{D}(\mathcal{R}(X))$, respectively). $\mathcal{D}(\mathcal{R}(X))$ shows to yield more distinguishable results, leading to a better model for one-class classification and hence video anomaly detection. It is fascinating to note that one of the most critical dilemmas for video anomaly detection methods is their high false positives. That is, algorithms often detect many of the `normal' scenes as anomalies. In Fig.  \ref{fig:ped2}, three left columns are three tough `normal' examples, as the human subject is not completely visible in the patch. We deliberately visualized these cases to illustrate how using $\mathcal{D}(\mathcal{R}(X))$ can effectively increase the discriminability of the system, compared to only $\mathcal{D}(X)$. 

\begin{table}[t]
\caption{Frame-level comparisons on Ped2}
\begin{center}
\begin{tabular}{ll|ll}
\hline
Method&EER&Method &EER    \\
\hline\hline
IBC~\cite{boiman2007detecting} & 13\%    & RE  \cite{sabokrou2016video} & 15\%  \\
MPCCA  ~\cite{kim2009observe} & 30\%   &  {\scriptsize Ravanbakhsh \etal}~\cite{ravanbakhsh2017training} &13\%\\
MDT ~\cite{mahadevan2010anomaly} & 24\% & {\scriptsize Ravanbakhsh \etal}~\cite{ravanbakhsh2017abnormal}& 14\%  \\
 {\scriptsize  Bertini \etal} ~\cite{bertini2012multi}& 30\% & {\scriptsize Dan Xu\etal}~\cite{xu2015learning}& 17\%  \\
  {\scriptsize Dan Xu \etal} ~\cite{xu2014video} &20\% &  {\scriptsize Sabokrou \etal}~\cite{sabokrou2015real} &19\% \\
Li \etal~\cite{li2014anomaly}&18.5\% &    Deep-cascade ~\cite{sabokrou2017deep} &  \textbf{9\%}   \\
\hline
Ours - $\mathcal{D}(X)$ & \textbf{16\%}  & Ours - $\mathcal{D}(\mathcal{R}(X))$ & \textbf{13\%}  \\
\hline
\end{tabular}
\end{center}
\label{tab:EER}
%NOT%\vspace{-2mm}
\end{table}

Similar to \cite{mahadevan2010anomaly}, we also report frame-level Equal Error Rate (EER) of our method and the compared methods. For this purpose, in any frame, if a pixel is detected as an anomaly, that frame is so labeled as `anomaly.' Table \ref{tab:EER} shows the result of our method in comparison to the state-of-the-art. The right column in Table \ref{tab:EER} lists the results from methods based on variations of deep-learning. This table confirms that our method is comparable to all these approaches, while we are solving a more general problem that can be used for any outlier, anomaly or novelty detection problem. It is worth noting that other methods (especially Deep-cascade ~\cite{sabokrou2017deep}) benefit from both spatial and temporal complex features, while our method operates on a patch-based basis, considering only spatial features of the frames. Our goal was to illustrate that the proposed method operates at least as well as the state-of-the-art, in a very general setting with no further tuning to the specific problem type. Simply, one can use spatiotemporal features and even further improve the results for anomaly event detection or related tasks.

\subsection{Discussion}
The experimental results confirmed that $\mathcal{R}$+$\mathcal{D}$ detects the novelty samples at least as well as the state-of-the-art or better than them in many cases, but finding the optimal structure and conducting the proper training procedure for these networks can be tedious and cumbersome tasks. The structure used in this paper proved well enough for our applications, while it can still be improved. We observed that achieving better performance is possible by modifying the structure, \eg, by some modification in the size and the order of convolutional layers of $\mathcal{R}$+$\mathcal{D}$, we achieved better results by margins of 0.02 to 0.04 compared to the results reported in Table \ref{tab:caltec}. Another important point is that it is very critical when to stop the training procedure of the joint network $\mathcal{R}$+$\mathcal{D}$. Stopping the training too early leads to immature learned network weights, while overtraining the networks confuses the $\mathcal{R}$ module and yields undesirable outputs. The stopping criterion outlined in Section \ref{sec:r+d} provides a right balance for the maturity of the joint network in understanding the underlying concept in the target class. 

In addition, it is important note is that training a model in absence of the novelty/outlier class can be considered as weak supervision. For many problems this is acceptable, as all the samples we have are often inliers. When dealing with outlier  detection problems, we can assume that number samples from the target class is much larger than the outlier samples. However, if we train the model at the presence of small number of outlier samples, the model still works. In a followup experiment, we mixed data from target (90\%) and outlier (10\%) classes in the training phase of the Ped2 experiment, and observed that the EER only dropped by 1.3\%, which is still comparable to the state-of-the-art methods. 

One of the major concerns in GANs is the mode collapse issue \cite{arjovsky2017wasserstein}, which often occurs when the generator only learns a portion of real-data distribution and outputs samples from a single mode (\ie, it ignores other modes). For our case, it is a different story as $\mathcal{R}$ directly sees all possible samples of the target class data and implicitly learns the manifold spanned by the target data distribution.
\section{Conclusion}

In this paper, we have proposed a general framework for one-class classification and novelty detection in images and videos, trained in an adversarial manner. Specifically, our architecture consists of two modules, $\mathcal{R}$econstructor and $\mathcal{D}$iscriminator. The former learns the concept of a target class to reconstruct images such that the latter is fooled to consider those reconstructed images as real target class images. After training the model, $\mathcal{R}$ can reconstruct target class samples correctly, while it distorts and decimates samples that do not have the concept shared among the target class samples. This eventually helps $\mathcal{D}$ discriminate the testing samples even better. We have used our models for a variety of related applications including outlier and anomaly detection in images and videos. The results on several datasets demonstrate that the proposed adversarially learned one-class classifier is capable of detecting samples not belonging to the target class (\ie, they are novelty, outliers or anomalies), even though there were no samples from the novelty class during training. 

\section*{Acknowledgement}
This research was in part supported by a grant from IPM (No. CS1396-5-01).

{\small
\bibliographystyle{ieee}
\bibliography{bibs}
}

\end{document}